\documentclass{article}

\usepackage{microtype}
\usepackage{graphicx}
\usepackage{subfigure}
\usepackage{booktabs} % for professional tables

\usepackage{amsmath}
\usepackage{amssymb}
\usepackage{mathtools}
\usepackage{amsthm}
\usepackage[section]{placeins}  % Float barrier

\usepackage{hyperref}

\usepackage[accepted]{icml2023}

\usepackage{amsmath}
\usepackage{amssymb}
\usepackage{mathtools}
\usepackage{amsthm}

\usepackage[capitalize,noabbrev]{cleveref}

\theoremstyle{plain}

\theoremstyle{definition}

\theoremstyle{remark}

\usepackage[textsize=tiny]{todonotes}

\icmltitlerunning{GraphGUIDE: interpretable and controllable conditional graph generation}

\begin{document}

\twocolumn[
\icmltitle{GraphGUIDE: interpretable and controllable conditional graph generation with discrete Bernoulli diffusion}



\icmlsetsymbol{equal}{*}

\begin{icmlauthorlist}
\icmlauthor{Alex M. Tseng}{genentech}
\icmlauthor{Nathaniel Diamant}{genentech}
\icmlauthor{Tommaso Biancalani}{genentech}
\icmlauthor{Gabriele Scalia}{genentech}
\end{icmlauthorlist}

\icmlaffiliation{genentech}{Department of Artificial Intelligence and Machine Learning, Research and Early Development, Genentech}

\icmlcorrespondingauthor{Alex M. Tseng}{tseng.alex@gene.com}

\icmlkeywords{diffusion model, generative model, conditional generation, graphs, molecules, interpretability}

\vskip 0.3in
]



\printAffiliationsAndNotice{}  % leave blank if no need to mention equal contribution

\begin{abstract}
Diffusion models achieve state-of-the-art performance in generating realistic objects and have been successfully applied to images, text, and videos. Recent work has shown that diffusion can also be defined on graphs, including graph representations of drug-like molecules. Unfortunately, it remains difficult to perform conditional generation on graphs in a way which is interpretable and controllable. In this work, we propose GraphGUIDE, a novel framework for graph generation using diffusion models, where edges in the graph are flipped or set at each discrete time step. We demonstrate GraphGUIDE on several graph datasets, and show that it enables full control over the conditional generation of arbitrary structural properties without relying on predefined labels. Our framework for graph diffusion can have a large impact on the interpretable conditional generation of graphs, including the generation of drug-like molecules with desired properties in a way which is informed by experimental evidence.
\end{abstract}

\section{Introduction}

Diffusion models have rapidly become the state-of-the-art method for generating many kinds of data \cite{Sohl-Dickstein2015,Ho2020,Song2021}, including images and videos \cite{Dhariwal2021,Ho2022}, text \cite{Li2022}, and tabular data \cite{Kotelnikov2022}. In order to generate data objects from some distribution $q_{0}(x)$, a diffusion model defines a \textit{forward-diffusion} process which iteratively adds noise (typically continuous Gaussian noise) to an object $x_{0} \sim q_{0}(x)$. The model then learns a \textit{reverse-diffusion} process to iteratively denoise a diffused object back to the original $x_{0}$. This allows the diffusion model to effectively sample new objects from $q_{0}(x)$. 

The most useful and impactful developments in diffusion models, however, arguably stem from the ability to \textit{conditionally} generate objects---that is, generating objects $x_{0}$ which satisfy some desired property. There are a few methods for conditional generation today, which effectively supply the model with a label $y$ (out of a predefined set) to steer the reverse-diffusion process toward generating an object $x_{0}$ which satisfies the property defined by $y$ (Section~\ref{sec:related}). Using these methods, many of the most prominent recent works in diffusion models have centered on conditional generation; these include image generation by class \cite{Song2021,Dhariwal2021} and image/video synthesis from text \cite{Rombach2021,Ho2022}. 

Although diffusion models have been successful in generating and conditionally generating many different data types, applying these models to generating \textit{graphs} poses several somewhat unique challenges, particularly for conditional generation. These challenges are major obstacles for many real-world problems, such as drug discovery, where it is critically important to be able to generate molecular graphs which have certain desired physiological or chemical properties. Although methods exist today for generating graphs using diffusion models \cite{Niu2020,Jo2022,Vignac2022}, they are limited in several ways.

One such limitation is that many current graph-diffusion methods are \textit{continuous} and generate reverse-diffusion intermediates which are not well-defined discrete graphs; instead, their intermediates include fractional or negative-valued edges. This severely hinders the interpretability and controllability of these methods; after all, it is not easy to decipher what $-0.2$ in an adjacency matrix means, or how to control generation where the probability of an edge is 0.6.

A bigger set of issues lies in the current available methods for \textit{conditional generation}, which are rather limiting for graph diffusion (including for diffusion frameworks which are discrete). Firstly, current conditional-generation methods supply a label $y$ to influence reverse diffusion as a soft constraint. This renders conditional generation an inscrutable process which prevents humans from interpreting or controlling the generated graph sample (this is even worse for continuous-diffusion frameworks). For many real-world applications such as drug discovery, being able to very precisely control the generated outputs for specific features (e.g. functional groups) is an important feature for a generative algorithm. Secondly, current conditional-generation methods rely on a predefined set of possible labels to condition on, and the addition of new properties necessitates the retraining of all or part of the model. For many graph applications, this is particularly limiting, as there are a huge number of specific structural properties that may be conditioned on. For example, in the fields of drug discovery and molecule design, it is not uncommon to desire a molecular graph which might contain a benzene, or toluene, or aniline, etc. Thus, for many real-world graph-generation problems such as drug design, both these limitations in current conditional-generation methods can pose a serious obstacle that prevents more widespread adoption of graph diffusion.

In this work, we present GraphGUIDE (\textbf{Graph} \textbf{G}eneration \textbf{U}sing \textbf{I}nterpretable \textbf{D}iffusion on \textbf{E}dges), an alternative graph-diffusion framework that addresses these limitations. GraphGUIDE relies on a diffusion process which is fully discrete, defined by flipping edges in and out of existence based on a Bernoulli random process. We define three different diffusion kernels, which add, delete, or flip edges randomly. Thus, at each point in both forward and reverse diffusion, the intermediate is an interpretable, well-defined graph. More importantly, this allows for \textit{full, interpretable control} of conditional generation using the appropriate kernel \textit{without any reliance on a predefined set of labels}.

To summarize, our main contributions are the following:

\begin{itemize}
    \item We present a novel framework, GraphGUIDE, for interpretable and controllable graph generation based on discrete diffusion of graph edges.
    \item We derive three discrete diffusion kernels based on the Bernoulli distribution, including a more efficient and stable parameterization of a kernel based on symmetrically flipping bits, and two novel kernels based on asymmetrically setting bits.
    \item We compare our method to other recent state-of-the-art graph-generation methods on benchmark datasets, showing that it achieves comparable generation quality. 
    \item We demonstrate that GraphGUIDE enables full control of arbitrary structural properties---which do not need to be predefined at training time---during generation, thus allowing the injection of custom priors such as the presence or absence of specific graph motifs.
\end{itemize}

\section{Related work}
\label{sec:related}

\subsection{Diffusion models}

\citet{Sohl-Dickstein2015} first described diffusion models as a method of generating data from some distribution. Given data samples $x_{0}$ drawn from some complex (and potentially high-dimensional) data distribution $q_{0}(x)$, the challenge of generative modeling is to find a way to model and generate novel data samples from the underlying distribution $q_{0}(x)$, even though it is intractable to describe and only a limited number of examples are available. To address this challenge, a diffusion model learns a \textit{bidirectional} mapping between the original data distribution $q_{0}(x)$ and some \textit{prior} distribution $\pi(x)$, where the prior distribution is tractable and can be sampled from easily. In a diffusion model, progressive amounts of random noise are added to some data sample $x_{0} \sim q_{0}(x)$ in the forward direction, thereby obtaining a noisy sample $x_{t} \sim q_{t}(x)$. As $t$ approaches the time horizon $T$ through many time steps, the distribution of noisy objects $q_{t}(x)$ approaches the tractable prior $q_{T}(x) = \pi(x)$. The core complexity of the diffusion model is to \textit{learn} the reverse-diffusion process by fitting a model $p_{\theta}(x_{t}, t)$ (typically a neural network) to effectively denoise any $x_{t}$ into a slightly less noisy $x_{t-1}$. After training the neural network on available data samples and many time points $t$, novel samples can be generated from $q_{0}(x)$ by first sampling from $\pi(x)$, and iteratively applying the neural network predictions to obtain progressively less and less noisy samples until reaching a final $x_{0} \sim q_{0}(x)$.

\subsection{Conditional generation}

Beyond the generation of samples from $q_{0}(x)$, a major goal of generative modeling is to perform \textit{conditional generation}, where we wish to draw a sample from $q_{0}(x)$ which satisfies a specific label or property. Within the diffusion-model literature, there are effectively two methods for conditional generation.

Classifier-guided conditional generation was proposed in \citet{Song2021}, in which an external classifier $f(x_{t})$ is trained on $x_{t}$ to predict some label $y$. Input gradients from this classifier are then used during the generative process to bias the generation of an object toward one which has the label $y$. While elegant in its mathematical justification (a simple invocation of Bayes' Rule), it relies on an external classifier which is trained on noisy inputs $x_{t}$ from across the diffusion timeline and a predefined set of labels $y$. This method is also only readily applied to diffusion models trained in a continuous-time and continuous-noise setting, due to its reliance on gradients.

In contrast to classifier-guided conditional generation, \citet{Ho2021} proposed an alternative method: classifier-free conditional generation. Instead of relying on an external classifier, the neural network $p_{\theta}$ (which defines the reverse-diffusion/generative process) is trained with labels as an input: $p_{\theta}(x_{t}, t, y)$. This method for conditional generation has been exceedingly popular, and has been shown to generate state-of-the-art samples \cite{Rombach2021,Ho2022}. Unlike classifier-guided conditional generation, this method enjoys the freedom of not relying on any external classifier, and it can be applied to discrete-time and/or discrete-noise diffusion settings.

Unfortunately, both methods for conditional generation suffer from some limitations. Firstly, both methods merely supply a reverse-diffusion-influencing signal to the generative process (i.e. through biasing gradients or through an auxiliary input). This acts as a \textit{soft constraint} which not only is uninterpretable, but also cannot be controlled or modified manually by a human during reverse diffusion. Secondly, both methods require a predefined set of labels $y$ (to train either an external classifier or the diffusion model itself). The addition of new labels or properties to conditionally generate would necessitate the retraining of an entire model; in the more popular case of classifier-free conditional generation, the entire diffusion model would need to be retrained.

\subsection{Discrete diffusion kernels}

The vast majority of diffusion models are trained with Gaussian noise, as this is the most well-developed kernel. Methods for discrete diffusion, however, have been proposed and utilized.

In the paper which first demonstrated diffusion models, \citet{Sohl-Dickstein2015} briefly proposed a discrete diffusion kernel based on the Bernoulli distribution---termed the ``binomial'' kernel---which flipped black-and-white pixels back and forth according to some probability. Unfortunately, the binomial kernel was parameterized in a way which made it inefficient for forward diffusion and unstable for reverse diffusion. Because of the large focus on Gaussian kernels in the literature, this discrete kernel has remained underdeveloped and unused until this point, and until now has not been reparameterized to behave efficiently and stably.

Later on, \citet{Hoogeboom2021} proposed an alternative way to diffuse over discrete states based on the multinomial distribution, where the forward-diffusion process slowly transforms a one-hot-encoded category vector into a uniform multinomial distribution. \citet{Austin2021} then demonstrated a similar framework where the forward-diffusion process is defined as Markov transitions over discrete states; the authors showed that multinomial diffusion is a special case of this framework. Although these methods for discrete diffusion have been successfully applied to unconditional generation, there is very limited usage of these kernels for conditional generation (in graphs or other data types).

\subsection{Graph generation}

For the problem of graph generation, there is already a large body of literature which spans many techniques, from autoregressive to one-shot. This includes GraphRNN \cite{You2018}, GRAN \cite{Liao2019}, MolGAN \cite{DeCao2018}, and SPECTRE \cite{Martinkus2022}, among others.

For graph generation using diffusion models specifically, a few methods have been proposed, each of which solves the problem of graph discreteness differently. \citet{Niu2020} demonstrated continuous diffusion on the adjacency matrix using Gaussian noise---a relaxation of the discreteness of edges. This generated graph-diffusion intermediates which had fractional and negative edges. \citet{Lee2022} then adapted this framework and performed conditional generation on molecular graphs by incorporating gradients from an external property-prediction network---an application of classifier guidance, made possible by the continuous relaxation \cite{Song2021}. Methods such as these, however---which diffuse on adjacency matrices---are severely hindered in their ability to conditionally generate structural properties. There are many equivalent adjacency-matrix orderings which satisfy the same structural property, which makes it difficult to inject this form of inductive bias into the generative process. Of course, these methods also suffer from the aforementioned limitations inherent to classifier-guided conditional generation.

Later on, \citet{Vignac2022} applied the Markovian diffusion framework proposed by \citet{Austin2021} on discrete graph adjacency matrices and node features. The authors showed that their method---DiGress---achieved state-of-the-art generative performance compared to other graph-generation methods, including those listed above. The authors also attempted to apply classifier-guided conditional generation using a discretized approximation of gradients, although this was limited by the conflict between continuous gradients and discrete diffusion, as well as the limitations inherent to current conditional-generation methods (i.e. they are uninterpretable soft constraints which require predefining a set of labels at training time).

\section{GraphGUIDE for conditional graph generation}

\begin{figure}[h]
\begin{center}
\centerline{\includegraphics[width=\columnwidth]{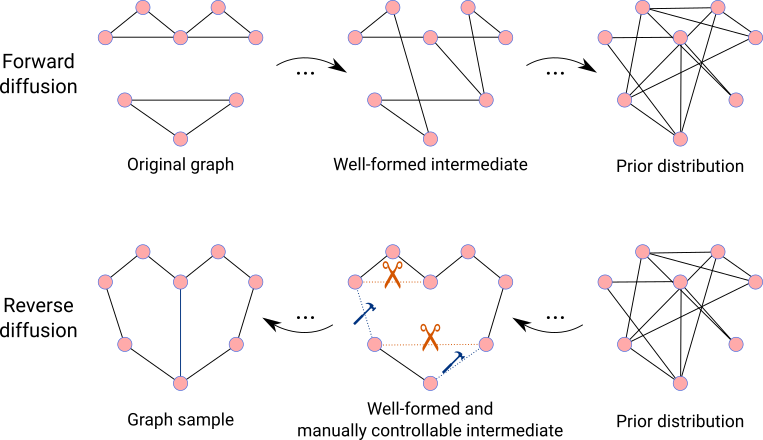}}
\caption{Interpretable and controllable graph diffusion. In order to generate discrete graphs using a diffusion model, we define the forward-diffusion noising process to flip edges in and out of existence to reach a prior distribution. All diffusion intermediates are interpretable well-defined graphs. In reverse diffusion, intermediates are also well formed, and therefore are fully manually controllable. In the GraphGUIDE framework, if a certain set of edges or graph motifs are desired (or undesired), appropriate edges may be manually added (or pruned, respectively) to generate a graph sample with the desired property.}
\label{fig:summary}
\end{center}
\end{figure}

Previous work in graph diffusion created uninterpretable diffusion intermediates, or they lacked a component which allowed for successful control over conditional generation. Here, we propose GraphGUIDE as an alternative framework for graph generation which allows for more interpretable and controllable generation (Figure \ref{fig:summary}).

In this section, we present discrete diffusion kernels for graph generation (Section~\ref{sec:bernoulli-kernels}), demonstrate that they are capable of achieving generative performance comparable to other state-of-the-art methods (Section~\ref{sec:perf}), and showcase the ease at which graph generation can be controlled for arbitrary structural properties using our GraphGUIDE framework (Section~\ref{sec:control}).

\subsection{Bernoulli diffusion on edges}
\label{sec:bernoulli-kernels}

\begin{figure}[h]
\begin{center}
\centerline{\includegraphics[width=\columnwidth]{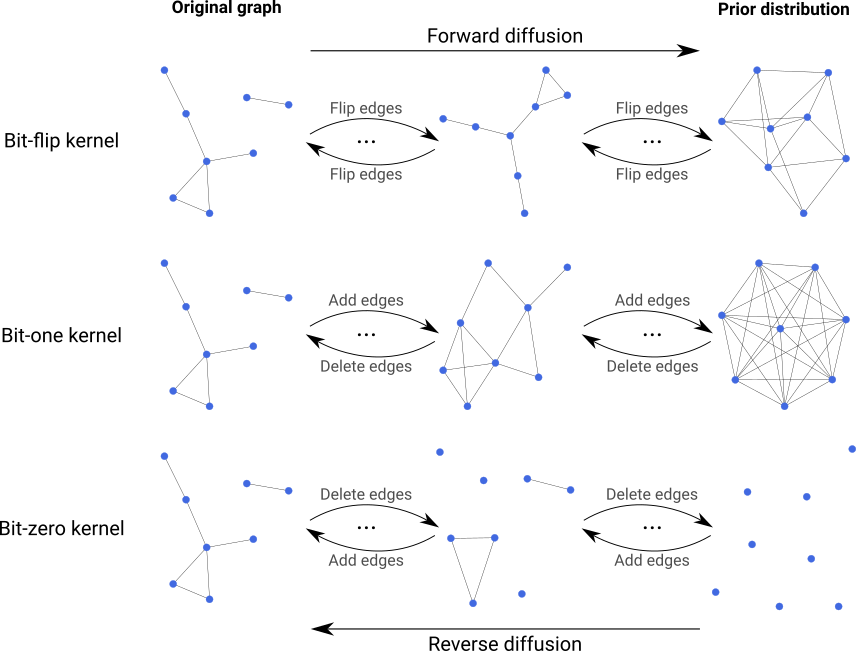}}
\caption{Summary of Bernoulli diffusion kernels applied to graph edges. The Bernoulli diffusion kernels operate on binary states. At each stage of forward diffusion, bits are flipped or set according to some probability. When applied to binary edge states in a graph (i.e. whether or not an edge exists), the Bernoulli kernels diffuse graphs by flipping (or setting) edges in or out of existence. This generates intermediates which are all well-defined graphs in both forward and reverse diffusion. We propose three Bernoulli diffusion kernels which flip edges on or off randomly (top), add edges randomly (middle), or delete edges randomly (bottom).}
\label{fig:kernel-summary}
\end{center}
\end{figure}

In order for GraphGUIDE to diffuse on graphs in a discrete and controllable manner, We define three discrete diffusion kernels based on the Bernoulli distribution (Figure~\ref{fig:kernel-summary}). Consider a binary vector $x_{t}$ to be diffused. The diffusion kernels will alter the bits randomly until a prior distribution is reached. Importantly, instead of performing diffusion on some continuous states, we define the Bernoulli diffusion kernels to \textit{directly} flip or set bits, thereby ensuring that every $x_{t}$ in the forward-diffusion process is a fully well-defined binary vector. We define three diffusion kernels for this process:

\begin{enumerate}
\item Bit-flip kernel: at time $t$, flip each bit with probability $\beta_{t}$.
\item Bit-one kernel: at time $t$, set each bit to 1 with probability $\beta_{t}$ (if the bit is already 1, do nothing).
\item Bit-zero kernel: at time $t$, set each bit to 0 with probability $\beta_{t}$ (if the bit is already 0, do nothing).
\end{enumerate}

For all three kernels, we assume there is a fixed noise schedule $\beta_{t}$ with $t \in \{1,...,T\}$. A typical noise schedule monotonically increases from $\beta_{1} = 0$ to $\beta_{T} = \frac{1}{2}$. Note that the forward-diffusion process defines an independent Bernoulli distribution for each entry in $x_{t}$ at every time $t$.

For each of these three kernels, we derive an analytical formula for the marginal forward distribution $q(x_{t}\vert x_{0})$ and for the marginal posterior distribution $q(x_{t-1}\vert x_{t},x_{0})$. Importantly, we derive parameterizations for these kernels which satisfy two properties: 1) $q(x_{t}\vert x_{0})$ is also a Bernoulli distribution for any $t$, and it is very tractable to compute its parameters; and 2) the reverse-diffusion posterior distribution $q(x_{t-1}\vert x_{t},x_{0})$ is also a tractable Bernoulli distribution, so the reverse-diffusion process can be modeled by learning the posterior's parameters directly, leading to more stable behavior. These two properties are critical for the efficient training of diffusion models which stably generate high-quality samples \cite{Ho2020}.

We present the formulae for our three Bernoulli kernels, including the forward distribution, the posterior distribution, and the prior distribution (Table~\ref{table:kernels}). See Appendix \ref{sec:kernels} for derivations.

\begin{table*}[t]
\caption{Forward, posterior, and prior distributions for three Bernoulli diffusion kernels. See Appendix \ref{sec:kernels} for derivations.}
\label{table:kernels}
\begin{center}
\resizebox{\textwidth}{!}{
\begin{tabular}{lp{0.3\linewidth}cc}
Kernel & Forward $q(x_{t} = 1\vert x_{0})$ & Posterior $q(x_{t-1} = 1\vert x_{t},x_{0})$ & Prior $\pi(x_{T} = 1)$ \\
\hline \\
Bit-flip & $(1-x_{0})\bar{\beta_{t}} + x_{0}(1-\bar{\beta_{t}})$,\newline where $\bar{\beta_{t}} = \frac{1}{2} - 2^{t-1}\prod\limits_{i=1}^{t}(\frac{1}{2}-\beta_{i})$ & $\frac{(x_{t} + \beta_{t} - 2x_{t}\beta_{t})(\frac{1}{2} + (2x_{0} - 1)2^{t-2}\prod\limits_{i=1}^{t-1}(\frac{1}{2}-\beta_{i}))}{\frac{1}{2} + (1-2(x_{t}-x_{0})^{2})2^{t-1}\prod\limits_{i=1}^{t}(\frac{1}{2}-\beta_{i})}$ & $\frac{1}{2}^{\dag}$ \\
Bit-one & $x_{0} + (1-x_{0})(1 - \prod\limits_{i=1}^{t}(1 - \beta_{i}))$ & $\frac{x_{t}(x_{0} + (1-x_{0})(1 - \prod\limits_{i=1}^{t-1}(1 - \beta_{i})))}{x_{0}x_{t} + (1-x_{0})x_{t}(1 - \prod\limits_{i=1}^{t}(1 - \beta_{i})) + (1-x_{0})(1-x_{t})\prod\limits_{i=1}^{t}(1-\beta_{i})}$ & $1^{\ddag}$ \\
Bit-zero & $x_{0}\prod\limits_{i=1}^{t}(1 - \beta_{i})$ & $\frac{(x_{t} + \beta_{t} - 2x_{t}\beta_{t})x_{0}\prod\limits_{i=1}^{t-1}(1 - \beta_{i})}{(1-x_{0})(1-x_{t}) + x_{0}(1-x_{t})(1 - \prod\limits_{i=1}^{t}(1 - \beta_{i})) + x_{0}x_{t}\prod\limits_{i=1}^{t}(1-\beta_{i})}$ & $0^{\ddag}$
\end{tabular}
}
\end{center}
\begin{small}
\dag Assuming $\beta_{t}$ approaches $\frac{1}{2}$ over a sufficiently long diffusion time\\
\ddag Assuming fractional $\beta_{t}$ over a sufficiently long diffusion time
\end{small}
\end{table*}

These three Bernoulli kernels are defined on general bits. In order to generate undirected graphs, we define $x_{t}$ as a binary vector denoting which edges exist in the graph (1 if the edge exists, and 0 otherwise). This vector has size $\binom{n}{2}$, where $n$ is the number of nodes in the graph (we do not allow self-edges or multi-edges, but our work can be extended easily to accommodate both cases; see Section~\ref{sec:discussion}). We diffuse on an undirected graph by applying a Bernoulli kernel to this binary edge vector. That is, the forward-diffusion process adds or removes edges (or both) randomly, and the reverse-diffusion process reconstructs a graph by deciding which nodes to link or unlink. The precise behavior depends on which of the three kernels is being used.

When applied to graph edges, the bit-flip kernel slowly flips edges in and out of existence with increasing probability until the graph approaches a prior which is the Erd\"os--Renyi graph (with $p = 0.5$ for a typical noise schedule). In reverse diffusion, the bit-flip kernel starts with an Erd\"os--Renyi graph and iteratively toggles edges on and off until a final graph sample is recovered. The bit-one kernel slowly adds edges randomly to the graph until reaching the prior, which is the complete graph (i.e. all possible edges exist). In the reverse direction, bit-one diffusion successively removes edges until obtaining a graph sample. Finally, the bit-zero kernel slowly removes random edges, approaching the prior of the empty graph (i.e. no edges exist at all). In reverse diffusion with the bit-zero kernel, edges are slowly added until a final graph sample is formed. In our experiments below, we focus our work on discrete diffusion over edge existence alone, as this is sufficient for many applications, including for molecule design (See Section~\ref{sec:discussion} for more details).

\subsection{Generative performance}
\label{sec:perf}

We compare the generative performance of graph-diffusion models trained with our Bernoulli kernels compared to other graph-generation methods. Over two well-known benchmark datasets (community-small and stochastic block models), we use the maximum-mean-discrepancy (MMD) metric to quantify how similar the generated graphs are to the training set (Tables \ref{table:performance-community}--\ref{table:performance-sbm}). We compute the MMD between the distributions of degrees, clustering coefficients, and orbit counts \cite{Hocevar2014}. We report an MMD ratio: the MMD of generated graphs and the training set, normalized by the MMD of the training set and an independently sampled validation set. We compare our MMD ratios to those reported by other graph-generation methods (when available). A lower MMD is better. Note that in some cases, these other methods erroneously label MMD \textit{squared} as MMD in their text, whereas we report MMD (i.e. values taken from other works which report MMD squared have been square rooted).

\begin{table}[ht]
\caption{Bernoulli edge diffusion MMD ratio (community-small)}
\label{table:performance-community}
\begin{center}
\begin{tabular}{lccc}
Method & Deg. $\downarrow$ & Clus. $\downarrow$ & Orbit $\downarrow$ \\
\hline \\
GraphRNN & 2.00 & 1.31 & 2.00 \\
GRAN & 1.73 & 1.25 & 1.00 \\
MolGAN & 1.73 & 1.36 & 1.00 \\
SPECTRE & 1.00 & 1.73 & 1.00 \\
DiGress & 1.00 & 0.95 & \textbf{1.00} \\
\hline \\
Bit-flip & \textbf{0.99} & \textbf{0.58} & 2.55 \\
Bit-one & 1.21 & 0.62 & 1.83 \\
Bit-zero & 1.87 & 1.02 & 4.69 \\
\end{tabular}
\end{center}
\end{table}

\begin{table}[ht]
\caption{Bernoulli edge diffusion MMD ratio (stochastic block models)}
\label{table:performance-sbm}
\begin{center}
\begin{tabular}{lccc}
Method & Deg. $\downarrow$ & Clus. $\downarrow$ & Orbit $\downarrow$ \\
\hline \\
GraphRNN & 2.62 & 1.33 & 1.75 \\
GRAN & 3.76 & 1.29 & 1.46 \\
MolGAN & 5.42 & 1.87 & 1.67 \\
SPECTRE & 3.14 & 1.26 & \textbf{0.54} \\
DiGress & 1.26 & 1.22 & 1.30 \\
\hline \\
Bit-flip & 2.73 & 1.23 & 0.94 \\
Bit-one & \textbf{1.00} & 1.21 & 0.81 \\
Bit-zero & 1.31 & \textbf{1.19} & 0.80 \\
\end{tabular}
\end{center}
\end{table}

These experiments demonstrate that diffusion using the discrete Bernoulli kernels on graph edges achieves comparable performance compared to other state-of-the-art methods for graph generation, including other discrete graph-diffusion methods such as DiGress \cite{Vignac2022}.

\subsection{Graph generation with interpretable control}
\label{sec:control}

GraphGUIDE employs these discrete Bernoulli kernels because they result in perfectly well-defined graphs at each intermediate stage of the diffusion process. Not only are these intermediates more readily interpretable, but they also allow the generation process to be easily controlled. At any stage of the reverse-diffusion process, edges or graph motifs that are desired can be manually retained in the intermediate (or symmetrically, edges or motifs that are not desired can be prevented from forming).

\begin{figure}[h]
\begin{center}
\centerline{\includegraphics[width=\columnwidth]{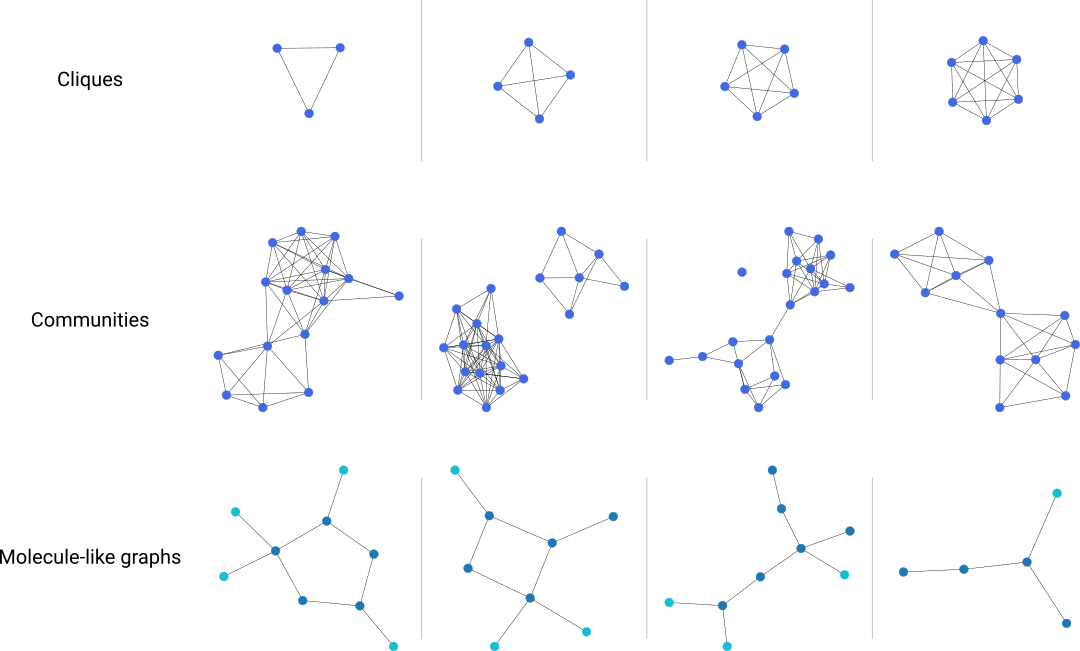}}
\caption{Examples of unconditionally generated graphs without manual control. We show graphs generated from a diffusion model trained on cliques of sizes ranging from 3 to 6, using the bit-one kernel (top). On the community-small dataset, we generated graphs from a diffusion model trained with the bit-zero kernel (middle). On a dataset of molecule-like graphs consisting of ring and non-ring backbones (dark blue) and secondary nodes (light blue), we generated graphs from a diffusion model trained with the bit-flip kernel (bottom).}
\label{fig:uncond}
\end{center}
\vskip -0.2in
\end{figure}

\begin{figure}[h]
\begin{center}
\centerline{\includegraphics[width=\columnwidth]{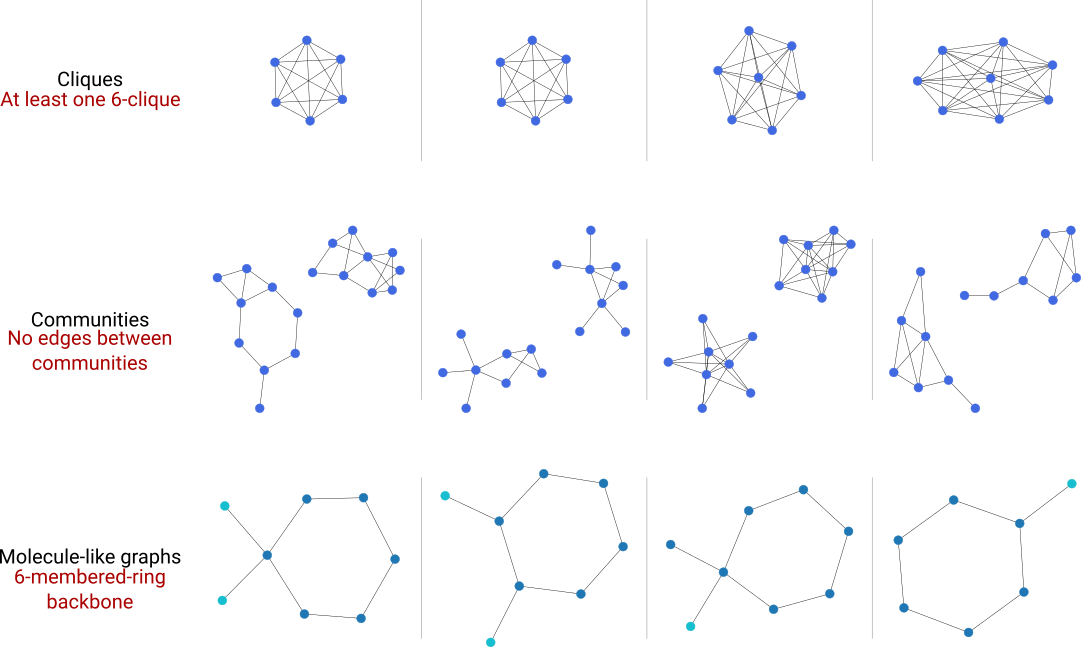}}
\caption{Examples of conditionally generated graphs through manual control with GraphGUIDE. On our model trained to generate cliques of various sizes using the bit-one kernel, we used manual control to enforce that the generated graph always had a 6-clique (top). On the community-small diffusion model (trained with the bit-zero kernel), we conditionally generated graphs by enforcing throughout the generation process that the two communities would remain disjoint (i.e. no edges between the communities) (middle). On the molecule-like diffusion model trained with the bit-flip kernel, we conditionally generated graphs such that the generated molecules always contained a 6-membered backbone ring (bottom).}
\label{fig:control}
\end{center}
\vskip -0.2in
\end{figure}

In order to illustrate the ease at which graph generation can be controlled with GraphGUIDE, we show some example graphs generated to have specific desired properties (Figures \ref{fig:uncond}--\ref{fig:control}). First, we trained a model using the bit-one kernel to generate cliques of sizes three to six. Graphs that were generated unconditionally (i.e. without manual control) contained cliques of various sizes (Figure~\ref{fig:uncond}, top). We then \textit{conditionally} generated graphs by manually controlling the reverse-diffusion process so that all generated graphs would have a clique of size 6 (Figure~\ref{fig:control}, top). Recall that the bit-one kernel confers a prior distribution which is the complete graph (i.e. $x_{T}$ corresponds to a graph with all possible edges). In the reverse-diffusion process, edges are gradually removed to recover a sample of $x_{0}$. For each graph generated under manual control, we arbitrarily selected 6 nodes. Throughout the reverse-diffusion process---at each step---we ensured that no edges were removed between any of these 6 nodes. If any such edges were removed at a reverse-diffusion step, they were added back before the next step. As a result, all graphs generated using this procedure contained a 6-clique. Intriguingly, the model also extrapolated from the data and generated 7- and 8-cliques, which contain 6-cliques as subgraphs.

We then trained a model using the bit-zero kernel to generate community-small graphs (as in Table \ref{table:performance-community}). Graphs generated by this model typically showed two communities which were oftentimes linked to each other with one or more edges (Figure~\ref{fig:uncond}, middle). We then controlled generation by ensuring no edges were added between the two communities, thereby always forming two disjoint subgraphs (Figure~\ref{fig:control}, middle). The bit-zero kernel confers a prior distribution which has no edges, and in the reverse-diffusion process, edges are slowly added to form the final graph sample $x_{0}$. In order to perform manual control, we simply partitioned the empty graph $x_{T}$ into two equal-sized sets of nodes, and ensured that no edges were ever added between nodes of different communities. The result is a set of graphs where the two communities were always disjoint.

Finally, we trained a model using the bit-flip kernel to generate a set of molecule-like graphs. Our molecule-like graphs consist of backbone and secondary nodes (atoms), where backbone nodes always have a degree of at most 4, and secondary nodes always have a degree of 1 and are linked to a backbone node. In terms of real organic chemistry, backbone nodes might correspond to carbon atoms, and secondary nodes might correspond to halogens. The backbone atoms can be linked together in rings of various sizes, or be acyclic and branched. Graphs generated from this model showed a diverse set of molecule-like structures (Figure~\ref{fig:uncond}, bottom). We then performed controllable generation and enforced throughout reverse diffusion that a 6-membered backbone ring would be formed (Figure \ref{fig:control}, bottom). The bit-flip kernel confers a prior distribution $x_{T}$ which is an Erd\"os--Renyi graph, and edges are slowly removed or added throughout the reverse-diffusion process to generate a graph sample $x_{0}$. In order to perform manual control, we identified 6 backbone nodes from $x_{T}$, and enforced throughout the generative process that those 6 backbone nodes were linked together in a ring, with no other edges between them. The resulting graphs always had a 6-membered backbone ring, usually with other backbone nodes or secondary nodes attached.

Throughout these experiments, because of the full manual control offered by the discrete diffusion, \textbf{100\%} of the resulting graphs always had the desired property. In contrast, when we unconditionally generated graphs from these models, we only obtained 6-cliques 68\% of the time, disjoint communities 21\% of the time, and 6-membered-ring molecules 12\% of the time.

In this section, we demonstrated the power of conditional generation via manual control with GraphGUIDE. Notably, this was made possible by the fully discrete nature of the Bernoulli kernels, which require that all diffusion intermediates are fully well-defined graphs. This is not only important for interpretability and for humans to easily manipulate the diffusion process, but is also critical for the robustness of the neural-network predictions. Just as the neural network is trained on well-defined graphs with binary edges, manually edited reverse-diffusion intermediates are also well-defined graphs with binary edges, and therefore are more likely in-distribution for the neural network, thus not leading to unexpected or undefined behavior. Furthermore, we presented \textit{three} Bernoulli kernels, each of which is well-suited for different controllable-generation tasks, where edges or motifs need to be retained or removed. The bit-one kernel is best for ensuring that a particular set of edges are retained; the bit-zero kernel is best for ensuring that a particular set of edges are removed; the bit-flip kernel is best for ensuring that a more complex motif (with some edges that need to be retained and others removed) forms. Using the best-suited kernel of the three helps ensure that manually controlled generation creates intermediates that are in-distribution for the model. For example, using the bit-one kernel generates a prior which is the complete graph, and so in the reverse-diffusion process, edges are gradually deleted. If the manual-control task is to retain a set of edges, then retaining those edges throughout the reverse-diffusion process ensures that diffusion intermediates are much more likely to remain in-distribution for the neural network.

Contrast this interpretability and controllability with other graph-diffusion frameworks. On continuous-diffusion frameworks, the diffusion intermediates contain edges that are fractional or negative. Not only is this far less interpretable, it is also much more difficult to control the generation in terms of desired structural features, such as the presence or absence of a particular motif. Any attempt to manually control the generative process like above would be foiled by the fact that fully well-defined graphs are not in-distribution for most of the diffusion process. Additionally, all existing graph-diffusion frameworks (continuous or discrete) rely on traditional methods for conditional generation, which inscrutably influence graph generation using a limited, predefined set of labels (Section \ref{sec:related}).

Our experiments highlight the unique method of conditional generation offered by GraphGUIDE. Conditional generation using GraphGUIDE no longer requires an external classifier, or even a predetermined set of property labels. Because the diffusion intermediates are so easily manipulated, practically any structural property---which is defined by the presence and/or absence of edges or motifs---can be imposed with full manual control at any time in the reverse-diffusion process. This very much distinguishes our framework from other methods for conditional generation.

\section{Discussion}
\label{sec:discussion}

The diffusion framework proposed in this work is unique from other diffusion methods due to the controllability of the reverse-diffusion product. Graph generation through GraphGUIDE can be thought of as a reverse-diffusion process which iteratively decides which edges to add or remove in order to recover a final graph sample $x_{0}$. This renders the generation process highly interpretable and controllable to humans.

Within the GraphGUIDE framework, nodes and their features can be thought of as static and eternal; the reverse-diffusion process determines which nodes to link up in order to form a final graph sample. As such, in our experiments, we trained our diffusion models to diffuse only on edges using our discrete Bernoulli kernels. That is, we did not perform diffusion on node features. This is because GraphGUIDE naturally defines reverse diffusion as a process which decides which nodes to link up with an edge. Of course, this requires the set of nodes to have a limited set of possible feature values. For many graph-generation tasks, this assumption holds, including graphs where only the structure is important (e.g. communities, stochastic block models, etc.), and molecular graphs (i.e. there are only a few types of atoms in typical molecular-generation tasks). Note, however, that in order to accommodate nodes with many more possible features, diffusion can also be performed on node features (which can be continuous or discrete) and on (discrete) edges jointly.

Although we demonstrated GraphGUIDE on undirected graphs without self-edges or multi-edges, our work can be easily extended to accommodate both, simply by having the binary vector $x_{t}$ (which denotes edge presence) contain an entry for every possible edge, whether it be a self-edge or a multi-edge (assuming there is a maximum number of multi-edges per pair of nodes). Similarly, this framework can also be readily applied to directed graphs by effectively doubling the size of $x_{t}$. Our method may also be applied to graphs with edges that have edge attributes, as long as there is a relatively small set of discrete attributes. Again, this can be accomplished by having additional entries in $x_{t}$, which denote every possible attribute for each edge. Alternatively, GraphGUIDE can be combined with kernels other than the Bernoulli kernels, such as multinomial kernels like those proposed in \citet{Hoogeboom2021} and \citet{Austin2021}. Graphs with many possible edge attributes or continuous edge attributes, however, do not fit very well into the GraphGUIDE framework, as a fundamental assumption made by the Bernoulli kernel (or multinomial kernels) is that all edges are binary (or have a limited number of discrete states, respectively).

Another limitation of our framework may be in the set of properties that can be conditionally generated via manual control. We demonstrated an exquisite degree of manual control afforded by discrete diffusion on edges, and this allowed conditional generation of graphs with arbitrary structural properties (e.g. molecules with 6-membered backbone rings), without the need to predefine them. Hence, in order to take advantage of controllable generation using our framework, the property desired must be definable structurally---that is, it needs to be definable by a set of known edges or graph motifs. As such, it is difficult to use this framework to generate graphs which satisfy a high-level property (e.g. a molecular graph which is a target for the $\beta_{2}$ adrenergic receptor), because it is not easy to identify what specific low-level structural properties (e.g. bonds or functional groups) confer this high-level property, which is a complex descriptor. Instead, high-level properties (e.g. drug targets) may be controlled for using the current standard methods of conditional generation, such as classifier-free conditional generation (i.e. training the diffusion model on a predefined set of labels) \citet{Ho2021}. Notably, our framework remains orthogonal to classifier-free conditional generation, and can be easily combined with such methods so that conditional generation can be performed on a smaller set of pre-defined high-level properties as well as arbitrary low-level structural properties simultaneously.

\section{Conclusion}

In this work, we proposed GraphGUIDE, a novel framework for interpretable and controllable graph generation using diffusion models. To aid in generating interpretable diffusion intermediates, we presented three discrete diffusion kernels based on the Bernoulli distribution and applied them to graph edges. The resulting diffusion processes add noise to graphs by flipping, adding, or removing edges until reaching some prior distribution. In the reverse-diffusion process, the diffusion model iteratively decides which edges to keep or remove to recover a final graph sample. Notably, all diffusion intermediates are fully well-defined graphs, thus allowing all diffusion intermediates to be interpretable. More importantly, by using the appropriate kernel, the generative process is highly controllable---specific edges, motifs, and other properties can be retained or prevented at each stage of the reverse-diffusion process, while still being in-distribution for the neural network. Because of this high degree of control over all parts of the generative process, practically any structural property can be conditioned on, and without relying on any predefined set of labels. Together, GraphGUIDE allows the enforcement of any arbitrary structural property on-the-fly with 100\% success. Additionally, we demonstrated that these advantages in interpretability and controllability are gained without any penalty in generative performance.

We illustrated the benefits of GraphGUIDE for several kinds of undirected graphs, particularly highlighting the application to molecular graphs for drug discovery or chemical design. Our work, however, may be applied to generating other kinds of graphs in an interpretable and controllable manner, such as knowledge graphs and causality graphs. In both these situations, it would be highly beneficial to be able to control for certain substructures easily and interpretably. Furthermore, the framework defined by GraphGUIDE may be applied to data types outside of graph edges, as well. Further exploration in discrete diffusion and controllable generation can continue to have impacts in many other real-world domains.

% \section*{Acknowledgements}

\clearpage

\bibliography{discrete_graph_diff}
\bibliographystyle{icml2023}

% Appendices

\newpage
\appendix
\onecolumn

\section{Derivation of Bernoulli kernels}
\label{sec:kernels}

For all three kernels, we assume there is a fixed noise schedule $\beta_{t}$ with $t \in \{1,...,T\}$. A typical noise schedule monotonically increases from $\beta_{1} = 0$ to $\beta_{T} = \frac{1}{2}$.

\subsection{Derivation of bit-flip kernel}

In order to ``diffuse'' or ``noise'' a binary vector $x_{0}$, we define a stochastic process in which bits are slowly flipped between 0 and 1. At time $t$, each bit independently flips with probability $\beta_{t}$, and is kept the same with probability $1-\beta_{t}$. Note that if $\beta_{t} = 0$, then no noising happens.

\textbf{Forward distribution}

Here, we seek a closed-form expression for $q(x_{t} = 1\vert x_{0})$ so that we may compute the forward distribution in an efficient way. We have that $x_{t} = b_{t} - x_{0}$, where $b_{t}$ is a binary vector denoting whether or not a flip (relative to $x_{0}$) happens over diffusion time from 0 to $t$.  From the piling-up lemma, we have that $b_{t} \sim Bern(\frac{1}{2} - 2^{t-1}\prod\limits_{i=1}^{t}(\frac{1}{2} - \beta_{i}))$.

This gives us a forward distribution that is also distributed according to a Bernoulli distribution, where $q(x_{t} = 1\vert x_{0}) = (1-x_{0})\bar{\beta_{t}} + x_{0}(1-\bar{\beta_{t}})$, where $\bar{\beta_{t}} = \frac{1}{2} - 2^{t-1}\prod\limits_{i=1}^{t}(\frac{1}{2}-\beta_{i})$

\textbf{Prior distribution}

$\lim\limits_{t\rightarrow T}\bar{\beta_{t}} = \frac{1}{2}$, as long as $T$ is a sufficiently large time horizon and $\beta_{t}$ approaches $\frac{1}{2}$. Thus, $q(x_{T} = 1) = \frac{1}{2}$, regardless of the value of $x_{0}$.

\textbf{Posterior distribution}

Here, we seek a closed-form expression for $q(x_{t-1} = 1\vert x_{t},x_{0})$ as a parameterized distribution. By Bayes' Rule, we have $q(x_{t-1}=1\vert x_{t},x_{0}) = \frac{q(x_{t}\vert x_{t-1}=1,x_{0})q(x_{t-1}=1\vert x_{0})}{q(x_{t}\vert x_{0})}$.

Let's look at each piece separately:

$q(x_{t}\vert x_{t-1}=1,x_{0}) = (1-x_{t})\beta_{t} + x_{t}(1-\beta_{t}) = x_{t} + \beta_{t} - 2x_{t}\beta_{t}$

$q(x_{t-1}=1\vert x_{0}) = (1-x_{0})(\frac{1}{2} - 2^{t-2}	\prod\limits_{i=1}^{t-1}(\frac{1}{2}-\beta_{i})) + x_{0}(\frac{1}{2} + 2^{t-2}	\prod\limits_{i=1}^{t-1}(\frac{1}{2}-\beta_{i})) = \frac{1}{2} + (2x_{0} - 1)2^{t-2}\prod\limits_{i=1}^{t-1}(\frac{1}{2}-\beta_{i})$

$q(x_{t}\vert x_{0}) = (x_{t}-x_{0})^{2}(\frac{1}{2} - 2^{t-1}	\prod\limits_{i=1}^{t}(\frac{1}{2}-\beta_{i})) + (1-(x_{t}-x_{0})^{2})(\frac{1}{2} + 2^{t-1}	\prod\limits_{i=1}^{t}(\frac{1}{2}-\beta_{i})) = \frac{1}{2} + (1-2(x_{t}-x_{0})^{2})2^{t-1}\prod\limits_{i=1}^{t}(\frac{1}{2}-\beta_{i})$

Note $(x_{t}-x_{0})^{2}$ is the xor between the two (binary) variables

That is, all together, we have that $x_{t-1} \sim Bern(\tilde{p}(x_{0},x_{t}))$, where $\tilde{p}(x_{0},x_{t}) = \frac{(x_{t} + \beta_{t} - 2x_{t}\beta_{t})(\frac{1}{2} + (2x_{0} - 1)2^{t-2}\prod\limits_{i=1}^{t-1}(\frac{1}{2}-\beta_{i}))}{\frac{1}{2} + (1-2(x_{t}-x_{0})^{2})2^{t-1}\prod\limits_{i=1}^{t}(\frac{1}{2}-\beta_{i})}$

\subsection{Derivation of bit-one kernel}

Here, we define a stochastic process on binary $x_{0}$ where bits are slowly turned into 1s. At time $t$, each bit independently turns into a 1 with probability $\beta_{t}$. If the bit already is 1, then nothing happens. Note that if $\beta_{t} = 0$, then no noising happens.

\textbf{Forward distribution}

Again, we seek a closed-form expression for $q(x_{t} = 1\vert x_{0})$ so that we may compute the forward distribution in an efficient way. We have that $q(x_{t}=1\vert x_{0}) = x_{0} + (1-x_{0})(1 - \prod\limits_{i=1}^{t}(1 - \beta_{i}))$ (i.e. if $x_{0} = 1$, then $x_{t} = 1$; otherwise, $x_{t}$ will be 1 if at least one of the $\beta_{i}$s is successful). Again, this is a Bernoulli distribution.

\textbf{Prior distribution}

$\lim\limits_{t\rightarrow T} \prod\limits_{i=1}^{t}(1 - \beta_{i}) = 0$, as long as $T$ is a sufficiently large time horizon and $\beta_{t}$ approaches some nonzero fraction (e.g. $\frac{1}{2}$). Thus, $q(x_{T} = 1) = 1$, regardless of the value of $x_{0}$.

\textbf{Posterior distribution}

We again start with Bayes' Rule: $q(x_{t-1}=1\vert x_{t},x_{0}) = \frac{q(x_{t}\vert x_{t-1}=1,x_{0})q(x_{t-1}=1\vert x_{0})}{q(x_{t}\vert x_{0})}$.

Let's look at each piece separately:

$q(x_{t}\vert x_{t-1}=1,x_{0}) = x_{t}$ (if $x_{t-1}=1$, then $x_{t} = 1$ has a probability of 1, $x_{t} = 0$ can’t happen and has a probability of 0)

$q(x_{t-1}=1\vert x_{0}) = x_{0} + (1-x_{0})(1 - \prod\limits_{i=1}^{t-1}(1 - \beta_{i}))$

$q(x_{t}\vert x_{0}) = x_{0}x_{t} + (1-x_{0})x_{t}(1 - \prod\limits_{i=1}^{t}(1 - \beta_{i})) + (1-x_{0})(1-x_{t})\prod\limits_{i=1}^{t}(1-\beta_{i})$

That is, all together, we have that $x_{t-1} \sim Bern(\tilde{p}(x_{0},x_{t}))$, where $\tilde{p}(x_{0},x_{t}) = \frac{x_{t}(x_{0} + (1-x_{0})(1 - \prod\limits_{i=1}^{t-1}(1 - \beta_{i})))}{x_{0}x_{t} + (1-x_{0})x_{t}(1 - \prod\limits_{i=1}^{t}(1 - \beta_{i})) + (1-x_{0})(1-x_{t})\prod\limits_{i=1}^{t}(1-\beta_{i})}$

\subsection{Derivation of bit-zero kernel}

Here, we define a stochastic process on binary $x_{0}$ where bits are slowly turned into 0s. At time $t$, each bit independently turns into a 0 with probability $\beta_{t}$. If the bit already is 0, then nothing happens. Note that if $\beta_{t} = 0$, then no noising happens.

\textbf{Forward distribution}

Again, we seek a closed-form expression for $q(x_{t} = 1\vert x_{0})$ so that we may compute the forward distribution in an efficient way. We have that $q(x_{t}=1\vert x_{0}) = x_{0}\prod\limits_{i=1}^{t}(1 - \beta_{i})$ (i.e. if $x_{0} = 0$, then $x_{t} = 0$; otherwise, $x_{t}$ will be 0 if none of the $\beta_{i}$s was successful). Again, this is a Bernoulli distribution.

\textbf{Prior distribution}

$\lim\limits_{t\rightarrow T} \prod\limits_{i=1}^{t}(1 - \beta_{i}) = 0$, as long as $T$ is a sufficiently large time horizon and $\beta_{t}$ approaches some nonzero fraction (e.g. $\frac{1}{2}$). Thus, $q(x_{T} = 1) = 0$, regardless of the value of $x_{0}$.

\textbf{Posterior distribution}

We again start with Bayes' Rule: $q(x_{t-1}=1\vert x_{t},x_{0}) = \frac{q(x_{t}\vert x_{t-1}=1,x_{0})q(x_{t-1}=1\vert x_{0})}{q(x_{t}\vert x_{0})}$.

Let's look at each piece separately:

$q(x_{t}\vert x_{t-1}=1,x_{0}) = x_{t}(1-\beta_{t}) + (1-x_{t})\beta_{t} = x_{t} + \beta_{t} - 2x_{t}\beta_{t}$ (if $x_{t-1}=1$, then $x_{t} = 1$ means $\beta_{t}$ failed; on the other hand, $x_{t} = 0$ means $\beta_{t}$ succeeded)

$q(x_{t-1}=1\vert x_{0}) = x_{0}\prod\limits_{i=1}^{t-1}(1 - \beta_{i})$

$q(x_{t}\vert x_{0}) = (1-x_{0})(1-x_{t}) + x_{0}(1-x_{t})(1 - \prod\limits_{i=1}^{t}(1 - \beta_{i})) + x_{0}x_{t}\prod\limits_{i=1}^{t}(1-\beta_{i})$

That is, all together, we have that $x_{t-1} \sim Bern(\tilde{p}(x_{0},x_{t}))$, where $\tilde{p}(x_{0},x_{t}) = \frac{(x_{t} + \beta_{t} - 2x_{t}\beta_{t})x_{0}\prod\limits_{i=1}^{t-1}(1 - \beta_{i})}{(1-x_{0})(1-x_{t}) + x_{0}(1-x_{t})(1 - \prod\limits_{i=1}^{t}(1 - \beta_{i})) + x_{0}x_{t}\prod\limits_{i=1}^{t}(1-\beta_{i})
}$

\clearpage

\subsection{Training/sampling algorithms}

\begin{algorithm}[h]
    \caption{Training with the Bernoulli kernels}
    \label{alg:train}
\begin{algorithmic}
    \STATE {\bfseries Input:} training set $\{x^{(k)}\}$
    \REPEAT
        \STATE Sample $x_{0}$ from training data $\{x^{(k)}\}$
        \STATE Sample $t \sim Unif(0, T)$
        \STATE Sample $x_{t} \sim q_{t}(x\vert x_{0})$
        \STATE Gradient descent on $p_{(\theta_{s},\theta_{i})}(x_{t}, t)$ to predict $x_{0}$
    \UNTIL convergence
\end{algorithmic}
\end{algorithm}

\begin{algorithm}[h]
    \caption{Sampling with the Bernoulli kernels}
    \label{alg:sample}
\begin{algorithmic}
    \STATE {\bfseries Input:} trained $p_{\theta}$
    \STATE Sample $x_{t} \leftarrow x_{T}$ from $\pi(x)$
    \FOR{$t = T$ to 0}
        \STATE $\hat{x_{0}} \leftarrow p_{\theta}(x_{t},t)$
        \STATE Sample $x_{t-1} \sim q(x_{t-1}\vert x_{t},\hat{x_{0}})$
        \STATE $x_{t}\leftarrow x_{t-1}$
    \ENDFOR
    \STATE Return $\hat{x}$
\end{algorithmic}
\end{algorithm}

\clearpage

\section{Supplementary Methods}

The code to generate the results and figures in this work is available at \texttt{https://github.com/Genentech/GraphGUIDE}.

We trained all of our models and performed all analyses on a single Nvidia Quadro P6000.

\subsection{Datasets}

\textbf{Community-small}

We generate a dataset of two-community graphs following the definition in \citet{You2018}. That is, we first pick $\vert V\vert \in [12, 20]$, which is split into two equal-sized communities (for an odd number, one community will have one more node than the other). Each community is generated by the Erd\"os--Renyi model with $p = 0.3$. We then add $0.05\vert V\vert$ edges uniformly between the two communities.

All nodes have a single feature with value 1.

We generate 6400 random graphs per epoch, where graphs across different epochs are generated independently.

For our generative-performance analyses, following \citet{You2018,Liao2019,Martinkus2022,Vignac2022}, we pre-generated and cached 200 random graphs, and trained only on these 200 repeatedly.

\textbf{Stochastic block models}

We generate a dataset of stochastic-block-model graphs following the definition in \citet{Martinkus2022}. That is, we first pick a number of blocks between $[2, 5]$. Each block size is independently and uniformly sampled from $[20, 40]$ nodes. The intra-block edge probability is 0.3, and inter-block edge probability is 0.05.

All nodes have a single feature with value 1.

For our generative-performance analyses, following \citet{You2018,Liao2019,Martinkus2022,Vignac2022}, we pre-generated and cached 200 random graphs, and trained only on these 200 repeatedly.

\textbf{Cliques}

We generate a dataset of clique graphs. Each graph has 10 nodes, and contains two cliques randomly drawn from sizes $\{3, 4, 5, 6\}$, without replacement. Excess nodes are left as singletons.

All nodes have a single feature with value 1.

We generate 6400 random graphs per epoch, where graphs across different epochs are generated independently.

\textbf{Molecule-like graphs}

We generate a dataset of molecule-like graphs. The number of backbone nodes is selected randomly from $[4, 6]$, and the number of secondary nodes is selected randomly from $[0, 4]$. The backbone nodes are first linked into a ring or a branched structure (with 0.5 probability of each). If the backbone is to be linked into a ring, it is linked into a single simple cycle. If the backbone is to be linked into a branched structure, a random tree is constructed such that the degree of each backbone node is at most 4. In order to do this, backbone nodes are iteratively and stochastically added to a structure which is slowly built up by randomly adding a leaf node to some pre-existing node with degree less than 4.

Once the backbone is constructed, secondary nodes are added uniformly at random to the backbone. Secondary nodes are always added as ``leaves'' (i.e. they always have degree 1, attached to a backbone node), such that the backbone nodes have degree at most 4. The random selection is done such that each free spot on a backbone node has equal probability of having a secondary node added.

Singleton backbone and secondary nodes are added (in equal amounts, or as equal as possible) to create a graph of 10 nodes.

All backbone nodes have a single feature with value 0, and all secondary nodes have a single feature with value 1.

We generate 6400 random graphs per epoch, where graphs across different epochs are generated independently.

\subsection{Model architecture}

For most of our experiments, we used a graph-attention network. We compute a time embedding as $[\sin(\frac{\pi}{2}\frac{t}{T}), \cos(\frac{\pi}{2}\frac{t}{T}), \frac{t}{T}]$. The time embedding is passed through a dense layer with 256 units. This output is concatenated with the node features. This is then passed through 2 dense layers of 256 units each, followed by a ReLU after each one.

We also compute a spectrum transformation matrix for each graph, where we take the eigendecomposition of the Laplacian $L = I - D^{\frac{1}{2}}AD^{\frac{1}{2}}$ for adjacency matrix $A$ and degree matrix $D$. We limit the eigenvectors to the 5 smallest eigenvalues (or the number of nodes in the graph if there are fewer than 5).

The node embeddings are passed through 5 GAT layers. In each layer, the input node embeddings are passed through a spectral convolution using the spectrum transformation matrix. The spectral convolution has the same number of units as the input node embeddings. This is concatenated with the original input node embeddings for the layer, and passed through a GAT with 8 attention heads of 32 units each. The GAT attends every node to every other, thereby ensuring that information is passed between all nodes even as edge connectivity changes over the course of diffusion. Edge presence/absence is encoded as a binary set of edge attributes which are passed to the GAT. The output of each GAT layer is passed through dropout of probability 0.1, a ReLU, and layer normalized with a summed residual connection with the GAT output. This is then passed twice through a series of dense units which maintain dimensionality, dropout, and layer normalization with a summed residual connection.

The final output of the GAT layers is passed through two dense layers of 256 units each (followed by a ReLU after each one), then a final dense layer which maps each node's embedding into a single scalar. The output is a probability of each edge, which is the product of the two endpoint nodes' scalar embeddings passed through a sigmoid.

For our benchmark of stochastic block models, we use the architecture in \citet{Vignac2022}.

Our loss function is the binary cross entropy between the model output and $x_{0}$ (which is binary).

\subsection{Training schedules}

For all discrete Bernoulli kernels, we used a noise schedule of $\beta_{t} = \min(\sigma(\frac{t}{100} - 10), \frac{1}{2}-10^{-6})$, where the $10^{-6}$ is for numerical stability. We set $T = 1000$ time steps.

For our experiments, we trained for 100 epochs with a learning rate of 0.001. We used a batch size of 32.

For our benchmarking of generative performance, we trained for 200 epochs with a learning rate of 0.001. We used a batch size of 32, except we used a batch size of 16 for the stochastic block models.

\subsection{Generative performance}

Following the convention set by \citet{You2018} and continued by other graph-generation works including \citet{Liao2019,Martinkus2022,Vignac2022}; etc., we computed various multivariate measurements on the generated graphs and the training graphs (i.e. node degrees, clustering coefficients, and node orbit counts), and compared these multivariate measurements using maximum mean discrepancy (MMD).

We computed node orbit counts using Orca \cite{Hocevar2014}.

Like previous works, we computed MMD using the Gaussian-total-variation kernel. For node degrees, we directly compared the count histograms; we used $\sigma = 1$ in the kernel. For clustering coefficients, we compared the histograms drawn with 100 bins; we used $\sigma = 0.1$ in the kernel. For node orbit counts, we computed the average orbit count for each orbit type, and directly compared the histograms with $\sigma = 30$ in the kernel. These settings match the settings published in previous works, including \citet{Martinkus2022}.

Following \citet{Martinkus2022} and \citet{Vignac2022}, we also computed the MMD between the training set and an equal-sized validation set drawn from the same distribution. We report the ratio of the MMDs as the MMD between generated and training graphs, divided by the MMD between training and validation graphs.

In order to compare our MMD ratios with other works, we took the published MMD values for GraphRNN, GRAN, MolGAN, and SPECTRE from \citet{Martinkus2022}. We took the MMD ratios for DiGress directly from \citet{Vignac2022}.

\textbf{Importantly, all cited previous works report MMD \textit{squared}, but the values are erroneously described as simply MMD in the text}. Here, we adjust for this error by simply taking the square root of the reported values before comparing them to ours.

We also note that significantly lower (by orders of magnitude) MMD values were achieved by training on only a small (e.g. 200) set of cached graphs, and computing the MMD of generated graphs to this training set. In contrast, the MMD was much larger if the training set consisted of graphs randomly generated on the fly (i.e. there is no caching of graphs), and if the generated graphs were compared to an independently sampled test set. Of course, in this work we trained and tested on the same small set of cached graphs in order to be comparable to previous methods. Unfortunately, however, this is less realistic and inherently suffers from train--test leakage. Instead, we propose that moving forward, generative models may be trained on larger samples and that MMD values are computed against an independently sampled test set.

\subsection{Controllable generation}

To generate cliques of size 6, we started with our model trained on the full cliques dataset, using the bit-one kernel. After sampling from the prior (i.e. the complete graph $K_{10}$), we selected 6 nodes arbitrarily. At each point in the reverse-diffusion process, we ensured that edges do not get removed between these 6 nodes. If a reverse-diffusion step removed any of those edges, they were added back before the next reverse-diffusion step (including the very last step which generates the final graph). In figures which show generated cliques, only one clique per graph is shown for clarity.

To generate disjoint communities, we started with our model trained on the full community-small dataset, using the bit-zero kernel. After sampling from the prior (i.e. the empty graph with size between 12 and 20), we arbitrarily partitioned the nodes into two communities of equal size (or as equal as possible in the case of an odd number of total nodes). At each point in the reverse-diffusion process, we ensured that edges do not get added between these the two communities. If a reverse-diffusion step added any of those edges, they were removed before the next reverse-diffusion step (including the very last step which generates the final graph). We consider a generated graph to have two distinct communities if the largest connected component has fewer than 60\% of the nodes.

To generate molecule-like graphs with a 6-membered backbone ring, we started with our model trained on the full molecule-like dataset, using the bit-flip kernel. After sampling from the prior (i.e. Erd\"os--Renyi graph of size 10 and $p = 0.5$, with an average of 6 backbone nodes), we arbitrarily selected 6 backbone nodes and assigned each a ring position. Any graphs without 6 backbone nodes were tossed out from the analysis. At each point in the reverse-diffusion process, we ensured that these 6 nodes were connected in a ring (according to the ring positioning), and had no other edges between them. If a reverse-diffusion step added or removed any improper edges, they were fixed before the next reverse-diffusion step (including the very last step which generates the final graph). In figures which show generated molecule-like graphs, excess singleton nodes were removed for clarity.

\end{document}